\documentclass[pdflatex,sn-mathphys-num]{sn-jnl}

\usepackage{graphicx}
\usepackage{multirow}
\usepackage{amsmath,amssymb,amsfonts}
\usepackage{amsthm}
\usepackage{mathrsfs}
\usepackage{xcolor}
\usepackage{textcomp}
\usepackage{manyfoot}
\usepackage{booktabs}
\usepackage{algorithm}
\usepackage{algorithmicx}
\usepackage{algpseudocode}
\usepackage{listings}
\usepackage{comment}
\usepackage{siunitx}
\usepackage{bbm}
\usepackage{etoolbox}
\usepackage{multicol}
\usepackage{tabularx}
\usepackage{svg}
\usepackage{float}
\usepackage[caption=false]{subfig}
\usepackage{pifont}
\usepackage{bm}

\hypersetup{hypertexnames=false}

\begin{document}

\title[Bionic Style Transfer for Humanoid Robots]{Bionic Human-Motion Style Transfer for Physically Executable Whole-Body Control of Humanoid Robots}

\author[1]{\fnm{Tianchen} \sur{Huang}}
\equalcont{These authors contributed equally to this work.}

\author[2]{\fnm{Mingkuan} \sur{Zhao}}
\equalcont{These authors contributed equally to this work.}

\author[1]{\fnm{Yang} \sur{Gao}}

\author[1]{\fnm{Feiyang} \sur{Yuan}}

\author[1]{\fnm{Junchi} \sur{Gu}}

\author[1]{\fnm{Xiaohu} \sur{Zhang}}

\author[3]{\fnm{Dongdong} \sur{Zhao}}

\author[3]{\fnm{Shi} \sur{Yan}}

\author*[1]{\fnm{Yu} \sur{Wang}}\email{wangyuustc@ustc.edu.cn}

\author*[1]{\fnm{Wei} \sur{Gao}}\email{weigao@ustc.edu.cn}

\author[1]{\fnm{Shiwu} \sur{Zhang}}

\affil[1]{\orgdiv{Institute of Humanoid Robots, Department of Precision Machinery and Precision Instrumentation}, \orgname{University of Science and Technology of China}, \orgaddress{\city{Hefei}, \state{Anhui}, \postcode{230026}, \country{China}}}

\affil[2]{\orgdiv{School of Computer Science and Technology, Faculty of Electronic and Information Engineering}, \orgname{Xi'an Jiaotong University}, \orgaddress{\city{Xi'an}, \state{Shaanxi}, \postcode{710049}, \country{China}}}

\affil[3]{\orgdiv{School of Information Science and Engineering}, \orgname{Lanzhou University}, \orgaddress{\city{Lanzhou}, \state{Gansu}, \postcode{730000}, \country{China}}}

\abstract{%
Expressive whole-body motion is important for humanoid robots operating in human environments, where robots are expected to move stably while presenting readable and adjustable body behaviors. However, most expressive motions are still obtained from fixed demonstrations or manually designed scripts, making it difficult to reuse a demonstrated style across different motion contents. Inspired by the way human motion styles convey affective and intentional cues through gait rhythm, posture, arm swing and body sway, this paper proposes a bionic generation-to-control framework for exemplar-driven style transfer on humanoid robots. Given a short human style exemplar and a target content motion, the proposed framework generates a stylized whole-body reference that preserves the intended motion content while transferring the demonstrated style. A physics-aware multi-condition latent diffusion model is developed to fuse style, content and trajectory conditions, and classifier-free guidance is used to adjust the style intensity without retraining. To improve hardware executability, contact-consistency and temporal-smoothness regularization are imposed on decoded motions during training. The generated references are then converted into G1-compatible robot references and executed by a preview-based whole-body tracking policy trained with a cluster-and-distill strategy. Simulation and Unitree G1 experiments show that the proposed method can transfer short human style exemplars to diverse robot motion contents, reduce contact and jitter artifacts compared with animation-oriented style-transfer baselines, and achieve a $96.0\%$ success rate over $125$ reported real-robot trials. The results demonstrate the feasibility of using short human motion exemplars as reusable bionic sources for physically executable expressive humanoid motion.
}

\keywords{bionic motion generation, humanoid robot, human motion style transfer, whole-body control, motion diffusion, physically executable motion, expressive locomotion}

\maketitle

\section{Introduction}
Humanoid robots are increasingly expected to work in social, service and human-coexistence environments. In these settings, stable locomotion and task completion are necessary but not sufficient: the robot's body motion should also be legible, human-like and adjustable. Human motion style is an important bionic source for this capability because people perceive affect and intent from gait rhythm, arm swing, trunk posture and movement dynamics~\cite{de2009bodies,zhang2020kinematic,schuetz2022embody,riemer2023emotionmotion}. Recent studies on robot emotion expression further indicate that body movement itself can shape how humans interpret a robot's state and communicative intent~\cite{matsumaru2022emotionalmovements,mahzoon2022vertical}. This makes expressive locomotion a relevant engineering target for humanoid robots, not only an animation problem.

Most humanoid control research still emphasizes stability margins, tracking errors, disturbance rejection and task success. These objectives are essential for real hardware, yet they usually treat a reference motion as fixed and pay limited attention to style variations on the same motion content. In contrast, arbitrary motion style transfer in computer animation can generate visually natural stylized motions from short exemplars~\cite{Aberman_2020,10.1109/MCG.2017.3271464,Jang_2022,2024Arbitrary}. Directly applying such generated sequences to humanoid robots is difficult because robot execution must satisfy contact stability, joint velocity limits, torque limits and closed-loop tracking constraints. A visually plausible motion may still contain stance-foot drift, ground penetration or high-frequency joint oscillations that make hardware execution unreliable.

This paper studies the following problem: given a content motion and a short human style exemplar, generate a full-body reference that preserves the content trajectory, expresses the target human motion style, and can be stably executed by a real humanoid robot. The content motion specifies task semantics such as walking direction or turning pattern; the style exemplar supplies bionic expressive cues such as slower rhythm, asymmetric sway or raised-arm posture; and the robot controller must turn the generated reference into physically executable joint commands.

This formulation introduces several coupled challenges. First, style amplification can alter contact rhythm and center-of-mass motion, which may reduce balance robustness. Second, diffusion-based generation may produce foot sliding, ground penetration and high-frequency joint jitter if contact and smoothness are only checked after sampling. Third, the whole-body tracker must handle different style intensities and motion contents without retraining for each generated sequence. Fourth, style recognizability and physical executability form a practical trade-off: stronger stylization may improve perceptual metrics but degrade support-foot stability or hardware tracking.

To address these issues, we propose a bionic generation-to-control pipeline for humanoid robots. The method is not intended as a pure animation generator or a pure controller trainer. Instead, it connects human-motion-inspired style transfer with robot execution through a physics-aware multi-condition latent diffusion model, retargeting and a preview-based whole-body tracking policy. Contact-consistency and temporal-smoothness regularization are computed on decoded motions during diffusion training, so that style cues are encouraged to remain within a robot-executable motion space before tracking. The tracker is trained with generated stylized trajectories and a cluster-and-distill procedure, improving robustness to style-dependent reference variations.

The main contributions are:
\begin{enumerate}
\item We formulate human-motion-inspired style transfer as a bionic whole-body control problem for humanoid robots, linking exemplar-based human motion styles with real robot executability.
\item We develop a physics-aware multi-condition latent diffusion model that injects style cues while suppressing contact inconsistency and high-frequency joint oscillations through decoded-motion regularization.
\item We integrate the generated stylized references with a whole-body tracking policy trained on style-diverse motion references through a cluster-and-distill strategy, enabling robust deployment of diverse stylized motions.
\item We validate the pipeline in simulation and on a Unitree G1 humanoid robot, reporting both perceptual style metrics and robot-oriented feasibility metrics.
\end{enumerate}

\section{Bionic Motivation and Related Background}
\label{sec:bionic_motivation}
\subsection{Human-Motion-Inspired Bionic Background}
Bionic robot research converts biological motion principles, morphology and interaction cues into engineered mechanisms and controllers. In humanoid and legged robots, human-inspired jumping, smooth foot trajectory planning, anthropomorphic exoskeleton design, and bio-inspired gait generation with CPGs or reflex mechanisms have been used to improve dynamic motion, physical feasibility and human--robot compatibility~\cite{wang2024bionicjumping,li2023realtimejbe,fang2025bioinspiredcpg}. These studies suggest that human motion should not be treated only as visual animation data. For humanoid robots, human-derived motion patterns must also be transformed into references that satisfy contact, balance and tracking constraints.

Classical humanoid locomotion control provides the physical substrate for this work. Preview control of ZMP~\cite{Kajita2003Preview}, capture-point and capturability analysis~\cite{Pratt2006CapturePoint,Koolen2012Capturability}, DCM/MPC formulations~\cite{Wieber2006MPC,Herdt2010MPC}, and optimization-based systems on full humanoids~\cite{Kuindersma2016Atlas} established how planned trajectories can respect balance and contact constraints. These methods usually focus on robust walking and task-space tracking rather than style variation. Our work uses this control perspective to constrain generated expressive motions toward a space that can be tracked by a humanoid platform.

\subsection{Expressive Body Motion as Style Source}
Human motion style is a perceptual and kinematic phenomenon. Studies in affective neuroscience and movement science show that emotions and intentions can be recognized from body expressions, gait, posture and motion dynamics~\cite{de2009bodies,zhang2020kinematic,schuetz2022embody,riemer2023emotionmotion}. For human--robot interaction, this matters because robots can communicate nonverbal states through body motion even when facial or speech channels are limited. Reviews and experiments on emotional robot movement report that gesture, body pose, vertical oscillation and movement timing influence perceived affect and clarity~\cite{matsumaru2022emotionalmovements,mahzoon2022vertical}. These findings support the bionic premise of this paper: human style exemplars can serve as compact sources of expressive cues for humanoid whole-body behaviors.

Compared with social-robot gesture studies, humanoid whole-body expression has a stronger physical constraint. A stylized gait modifies support timing, trunk posture and arm swing simultaneously. Therefore, expressive locomotion cannot be treated as an upper-body animation overlay; it must be designed together with contact consistency, balance and tracking robustness.

\subsection{Robot Executability Gap in Generative Style Transfer}
Motion style transfer aims to change the style of a motion while preserving its content. Feed-forward and unpaired methods demonstrated that content and style can be separated and recombined for character animation~\cite{10.1109/MCG.2017.3271464,Aberman_2020,2020Unpaired}. Motion Puzzle introduced body-part-level composition for arbitrary style transfer~\cite{Jang_2022}, and recent latent or diffusion-based methods further improved diversity and controllability~\cite{2024MoST,2024Generative,2024Arbitrary,zhong2024smoodi,hu2024diffusionbasedhumanmotionstyle}. Text-to-motion diffusion models such as MDM, MotionDiffuse and MoFusion also provide a broader foundation for conditional human motion generation~\cite{tevet2022,zhang2022motion,dabral2022mofusion}.

These animation-oriented methods are important baselines, but they rarely guarantee robot executability. A generated motion may look natural or carry recognizable style cues in kinematic evaluation, while still exhibiting stance-foot sliding, ground penetration or high-frequency joint jitter that violates hardware constraints. Physics-guided motion generation methods such as PhysDiff identify floating, foot sliding and penetration as common artifacts in motion diffusion~\cite{yuan2023physdiff}. Our method follows this deployment-oriented motivation, but focuses on exemplar-driven style transfer for humanoid robots: physics-aware losses are imposed on decoded stylized references, and the resulting motions are further retargeted and tracked by a whole-body controller.

\subsection{Whole-Body Control Requirements for Humanoid Execution}
Whole-body tracking has become a practical route for deploying diverse humanoid skills. DeepMimic and AMP showed how reference tracking and adversarial motion priors can produce natural-looking simulated skills~\cite{Peng_2018,Peng_2021}. PHC and universal humanoid representations improved persistent imitation and recovery across large motion sets~\cite{luo2023phc,luo2024}. Recent humanoid trackers further emphasize deployment: ExBody and ExBody2 focus on expressive upper-body imitation with robust locomotion~\cite{cheng2024,ji2025exbody2}, GMT and Any2Track pursue generalized tracking under motion diversity and disturbances~\cite{chen2025gmt,zhang2025trackmotionsdisturbances}, and ASAP, BeyondMimic, KungfuBot and related systems address sim-to-real adaptation or highly dynamic skills~\cite{he2025asap,liao2025beyondmimic,xie2025kungfubot,han2025kungfubot2}.

Our work is complementary to these trackers. Rather than only asking whether a policy can track an existing reference, we study how to generate a short-exemplar-based stylized reference that remains suitable for tracking. The closed loop from human style exemplar to physics-aware generation, retargeting, tracking and real-robot validation is the central distinction.

\section{Human-Motion-Inspired Style Generation}
\label{sec:style_generation}

\subsection{Pipeline Overview}
A unified pipeline from human-motion style generation to humanoid execution is presented. For any executable content motion from MoCap, video-derived skeletons or open motion repositories, the user supplies a short human style exemplar together with a guidance scale $\lambda$ to control style intensity at inference. The content motion specifies the intended action and global trajectory, such as straight walking, turning or circular walking. The style exemplar provides reusable human-motion cues, such as gait rhythm, body sway, arm posture and asymmetric movement tendency. The goal is to transfer the demonstrated style to the given content motion, instead of replaying the style exemplar as a fixed scripted behavior.

The pipeline synthesizes a stylized motion using a physics-aware multi-condition latent diffusion model, converts the generated motion into a G1-compatible reference, and executes it on the robot with a whole-body tracking policy. The generation model is conditioned on content, style and trajectory so that the produced motion can preserve the target action, express the human-derived style and maintain the global path structure. Contact-consistency and temporal-smoothness regularization are applied during generation training to reduce artifacts that would hinder downstream execution. The tracker is trained with both human motion data and generated stylized trajectories, allowing it to adapt to style-dependent reference variations without switching controllers. This cascaded design separates bionic style transfer at the generation stage from physical execution at the tracking stage, and yields style-faithful whole-body behaviors on humanoid robots.

\begin{figure}[t]
    \centering
    \includegraphics[width=\textwidth]{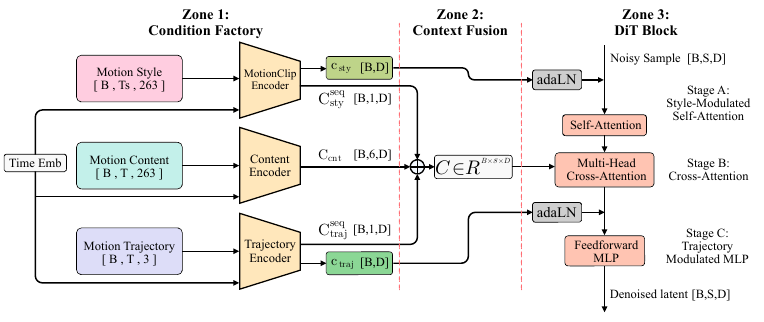}
    \caption{Physics-aware multi-condition latent diffusion architecture. Content, style and trajectory streams are encoded and fused as conditional context for the denoising Transformer. The style stream transfers human-derived motion cues from a short exemplar, the content stream preserves the target action, and the trajectory stream maintains the global path. Contact-consistency and temporal-smoothness losses are computed on decoded motions during training to improve downstream humanoid execution.}
    \label{fig:mcd}
\end{figure}

\subsection{Physics-Aware Multi-Condition Latent Diffusion}
\label{sec:diffusion}

As illustrated in Fig.~\ref{fig:mcd}, the motion generation framework follows a multi-condition latent diffusion formulation for exemplar-driven motion style transfer. Similar to MCM-LDM~\cite{2024Arbitrary}, content, style and trajectory are used as complementary conditions during denoising. The difference lies in how these conditions are fused and optimized for humanoid execution. Instead of relying only on content concatenation and adaptive modulation of self-attention and feed-forward layers, the proposed denoiser organizes content, style and trajectory as explicit condition tokens and fuses them with motion latent tokens through cross-attention.

The framework consists of multi-stream condition encoding, context fusion and a Transformer denoiser with staged conditioning. In each denoising block, the style condition modulates the self-attention branch, the fused condition context is used for cross-attention, and the trajectory condition modulates the feed-forward branch. This design allows motion tokens to retrieve content, style and trajectory cues at the token level, while the subsequent physics-aware regularization constrains the generated references toward downstream humanoid execution.

\subsubsection{Framework Architecture}

The diffusion model is defined in the latent space of a pre-trained motion VAE. For each motion sequence, the VAE encoder maps the input motion to a fixed-length latent token sequence. At timestep $\tau$, the denoiser takes as input the noised latent token sequence $\mathbf{x}_{\tau}\in\mathbb{R}^{B\times S\times D}$ together with three conditional streams, namely content, style and trajectory, and predicts the additive Gaussian noise. Here, $B$, $S$ and $D$ denote the batch size, the number of latent tokens and the latent dimension, respectively. The reverse diffusion process estimates clean latent tokens in the VAE latent space. These denoised latent tokens are then decoded by the frozen VAE decoder to recover motion features, which are subsequently used for retargeting and tracking. To synchronize the three conditional streams, the timestep $\tau$ is encoded by a sinusoidal embedding followed by an MLP to obtain a diffusion-step embedding $e_{\tau}\in\mathbb{R}^{B\times D}$, which is injected additively into the three condition representations before they are fused.

For the content stream, the root translation of the content motion $x_{\mathrm{cnt}}$ is removed before being fed into the content encoder. The encoder is pre-trained with SmoothL1 reconstruction in feature and joint spaces and a KL regularizer as
\begin{equation}
\begin{split}
\mathcal{L}_{\mathrm{vae}} &= \beta_{\mathrm{rec}}\|x-\hat{x}\|_{\mathrm{SL1}} 
+ \beta_{\mathrm{joint}}\|\mathrm{J}(x)-\mathrm{J}(\hat{x})\|_{\mathrm{SL1}} \\
&\quad + \beta_{\mathrm{kl}} \mathrm{KL}\left(q_{\phi}(z|x)\,\|\,\mathcal{N}(0,I)\right),
\end{split}
\label{eq:vae_loss}
\end{equation}
where $x\in\mathbb{R}^{B\times T\times 263}$ is the input motion feature sequence, $\mathrm{J}(\cdot)$ is the deterministic mapping to joint space, $q_{\phi}(z|x)$ is a Gaussian posterior parameterized over a sequence of latent tokens, $\hat{x}$ is the motion feature reconstructed by the decoder, and $\beta_{\mathrm{rec}}$, $\beta_{\mathrm{joint}}$ and $\beta_{\mathrm{kl}}$ are VAE pre-training weights. After pre-training, both the VAE encoder and decoder are kept fixed during diffusion training and inference. The produced latent token sequence is instance-normalized, shifted by $e_{\tau}$ through broadcasted addition, augmented with positional encoding, and refined by a single-layer Transformer encoder. A learned token-mixing linear projection is then used to obtain the final content token sequence $C_{\mathrm{cnt}}\in\mathbb{R}^{B\times 6\times D}$.

For the trajectory stream, the root translation sequence $c_{\mathrm{path}}\in\mathbb{R}^{B\times T\times 3}$ extracted from the content motion is encoded by a Transformer-based trajectory encoder. The output single global trajectory token is shifted by $e_{\tau}$ to obtain the time-conditioned trajectory token sequence $C_{\mathrm{traj}}^{\mathrm{seq}}\in\mathbb{R}^{B\times 1\times D}$, and its squeezed version defines the global trajectory vector $c_{\mathrm{traj}}\in\mathbb{R}^{B\times D}$.

For the style stream, the style motion $x_{\mathrm{sty}}$ is encoded by a frozen MotionCLIP encoder~\cite{tevet2022motionclip} into a single embedding vector. This embedding is projected to the diffusion latent dimension by a ReLU-linear layer and shifted by the timestep embedding $e_{\tau}$ to form a time-conditioned single-token sequence $C_{\mathrm{sty}}^{\mathrm{seq}}\in\mathbb{R}^{B\times 1\times D}$. The same tensor with the token dimension removed defines the global style vector $c_{\mathrm{sty}}\in\mathbb{R}^{B\times D}$, which is used to condition the adaptive normalization parameters of the self-attention branch in each denoiser block.

Condition fusion is implemented inside a multi-condition Diffusion Transformer. The three time-conditioned token sequences are concatenated along the sequence dimension to form an explicit condition context,
\begin{equation}
C=[C_{\mathrm{cnt}};\,C_{\mathrm{sty}}^{\mathrm{seq}};\,C_{\mathrm{traj}}^{\mathrm{seq}}]\in\mathbb{R}^{B\times 8\times D}.
\end{equation}
Different from modulation-only fusion used in prior multi-condition style-transfer diffusion models~\cite{2024Arbitrary}, this context is used as Keys and Values in a cross-attention branch, while the current motion latent tokens serve as Queries. This enables the denoising tokens to retrieve content, style and trajectory information directly from the condition tokens at each denoising stage.

Each denoiser block contains a style-conditioned self-attention branch, a cross-attention branch over $C$, and a trajectory-conditioned feed-forward branch. In the self-attention branch, LayerNorm is applied without affine parameters, and the affine shift, affine scale and residual gate are produced by an MLP conditioned on $c_{\mathrm{sty}}$. In the feed-forward branch, an MLP conditioned on $c_{\mathrm{traj}}$ produces the corresponding affine and gating parameters. The cross-attention branch adds the context-dependent update to the motion tokens through a residual connection.

\subsubsection{Training Objective and Controllable Inference}

Given a clean latent motion $\mathbf{x}_0$, the forward process of the conditional latent diffusion produces $\mathbf{x}_{\tau}$ by injecting Gaussian noise under a fixed schedule. The denoiser $\epsilon_{\theta}$ is trained with a mean squared error objective as
\begin{equation}
\mathcal{L}_{\mathrm{diff}} = \mathbb{E}_{\mathbf{x}_0, \tau, \epsilon} \left[ \left\| \epsilon - \epsilon_{\theta}(\mathbf{x}_{\tau}, \tau, C, c_{\mathrm{sty}}, c_{\mathrm{traj}}) \right\|_2^2 \right].
\label{eq:diff_loss}
\end{equation}
Classifier-free guidance is enabled by randomly dropping the style condition with probability $p_{\mathrm{drop}}$ during training. Specifically, the MotionCLIP style embedding before the style projection network is replaced by a zero vector for unconditional noise prediction, while the content and trajectory conditions are kept unchanged. During inference, the unconditional and conditional noise predictions are computed and combined using a guidance scale $\lambda$ as
\begin{equation}
\hat{\epsilon}_{\tau}=\epsilon_{\theta}^{(u)}+\lambda\left[\epsilon_{\theta}^{(c)}-\epsilon_{\theta}^{(u)}\right],
\end{equation}
where $\epsilon_{\theta}^{(u)}$ denotes the denoiser output obtained by using a zero style encoding, and $\epsilon_{\theta}^{(c)}$ denotes the output obtained with the original style encoding. The guidance scale $\lambda$ therefore controls the strength of the transferred style at inference without retraining the generation model. After iterative denoising, the final latent estimate is decoded by the frozen VAE decoder to produce the stylized motion feature sequence.

\subsubsection{Physics-Aware Regularization}

While latent diffusion models can generate diverse stylized motions, kinematically plausible results are not necessarily executable by a humanoid robot. In contrast to animation-oriented style-transfer objectives that mainly evaluate content preservation, style expression and trajectory consistency, we augment the diffusion objective with physics-aware regularization for robot deployment. These terms are not computed directly on $\mathbf{x}_{\tau}$, but on joint trajectories decoded from the model's estimate of the clean latent motion at a randomly sampled training timestep. This design biases the denoising model toward references with fewer contact and smoothness artifacts before they are converted to G1-compatible references and tracked.

Specifically, for each training sample, $\mathbf{x}_{\tau}$ is formed by adding Gaussian noise to the VAE latent tokens, and a noise prediction $\hat{\epsilon}_{\tau}$ is obtained. Using the noise-scheduler coefficient $\bar{\alpha}_{\tau}$, an estimate of the clean latent tokens is constructed as
\begin{equation}
\hat{\mathbf{x}}_0=\frac{\mathbf{x}_{\tau}-\sqrt{1-\bar{\alpha}_{\tau}}\,\hat{\epsilon}_{\tau}}{\sqrt{\bar{\alpha}_{\tau}}}.
\end{equation}
$\hat{\mathbf{x}}_0$ is then decoded with the frozen VAE decoder to obtain predicted motion features, which are converted to joint positions using the mapping function $\mathrm{J}(\cdot)$. The corresponding joint references are obtained by applying the same mapping to the ground-truth motion features in the training batch. This results in the predicted and reference joint-position trajectories $(q^{\mathrm{pred}},q^{\mathrm{ref}})$, where $q_{f,t}$ denotes the position of foot-related joint $f$ at frame $t$. These trajectories are used only during training for the two losses below.

First, to enforce contact stability, undesired foot sliding and ground penetration are penalized during stance. Let $\mathcal{F}$ denote the foot-related joints used in the loss. The reference foot speed is defined by finite differences of joint positions as $v^{\mathrm{ref}}_{f,t}=\|q^{\mathrm{ref}}_{f,t}-q^{\mathrm{ref}}_{f,t-1}\|_2$, and the stance mask is constructed as $s_{f,t}=\mathbbm{1}[v^{\mathrm{ref}}_{f,t}<\delta_v]$ with $\delta_v=0.002$. The contact loss applies hinge penalties to the predicted foot speed $v^{\mathrm{pred}}_{f,t}=\|q^{\mathrm{pred}}_{f,t}-q^{\mathrm{pred}}_{f,t-1}\|_2$ and height $h^{\mathrm{pred}}_{f,t}$ with a penetration tolerance $\delta_h=0.001$ as
\begin{equation}
\begin{split}
\mathcal{L}_{\mathrm{contact}} = \frac{1}{|\mathcal{F}|(T-1)} & \sum_{f \in \mathcal{F}} \sum_{t=2}^{T} s_{f,t} \Big[ \mathrm{ReLU}(v^{\mathrm{pred}}_{f,t}-\delta_v) \\
& + \mathrm{ReLU}(-h^{\mathrm{pred}}_{f,t}-\delta_h) \Big].
\end{split}
\label{eq:contact_loss}
\end{equation}

Second, to suppress high-frequency joint oscillations, a smoothness regularization is imposed on first- and second-order temporal differences of the joint trajectories. Let $\mathbf{v}_t=q_t-q_{t-1}$ and $\mathbf{a}_t=\mathbf{v}_t-\mathbf{v}_{t-1}$ denote the velocity and acceleration tensors. The smoothness loss matches predicted and reference derivatives using $\ell_1$ penalties as
\begin{equation}
\mathcal{L}_{\mathrm{smooth}} = \left\| \mathbf{v}^{\mathrm{pred}} - \mathbf{v}^{\mathrm{ref}} \right\|_1 + \left\| \mathbf{a}^{\mathrm{pred}} - \mathbf{a}^{\mathrm{ref}} \right\|_1.
\label{eq:smooth_loss}
\end{equation}
Both $\mathcal{L}_{\mathrm{contact}}$ and $\mathcal{L}_{\mathrm{smooth}}$ require reference motions to construct $q^{\mathrm{ref}}$ and the stance mask, and are therefore applied only during training.

\section{Humanoid Control and Deployment}
\label{sec:humanoid_control}

\subsection{Reference-Preview Whole-Body Tracking}
Before tracking, each generated motion is converted offline into a G1-compatible reference with joint ordering, joint-limit handling and kinematic filtering. The same conversion procedure is used for all generated motions and all baseline motions before robot-oriented evaluation.

Whole-body motion tracking is then formulated as closed-loop control with respect to the converted reference. The policy receives proprioceptive observations and a short-horizon future reference. The proprioceptive observation includes body orientation, body angular velocity, joint positions, joint velocities, the previous action and a fixed-length history buffer. The future reference covers a one-second preview window with $10$ uniformly sampled future instants. For each preview instant, root height, roll and pitch, root linear velocity, yaw angular velocity in the local root frame, and joint references are provided.

The preview features are concatenated into a temporal feature, compressed by a lightweight temporal encoder, and fused with the proprioceptive observation. This design allows the policy to anticipate style-dependent rhythm and contact timing, which is important for stabilizing the robot near swing--stance transitions. The policy outputs target joint positions at $50$ Hz, which are tracked by a low-level PD controller running at $1000$ Hz. Joint-specific PD gains are used on the real robot, with proportional gains ranging from $50$ to $200$ across different joints.

\subsection{Style-Diverse Cluster-and-Distill Training}
The tracker is trained to handle the G1-compatible reference distribution produced from the proposed style-transfer model. The training set contains G1-compatible references converted from LaFAN1, a filtered subset of AMASS, and $1000$ stylized trajectories generated by the proposed generator. In this work, we retain locomotion-related and whole-body motions that can be retargeted to the G1 platform, resulting in $77$ LaFAN1 sequences and $3142$ AMASS clips for tracker training. Non-executable motions are filtered following the PHC protocol~\cite{luo2023phc}. Introducing stylized trajectories allows the policy to encounter and adapt to them from the outset of training, because style transfer can change gait rhythm, trunk motion, arm posture and contact timing.

The expert policies are trained with a DeepMimic-style reference-tracking reward, following the design principle used in recent humanoid motion-tracking systems: reference-tracking terms are kept positive and dominant, while a small number of regularization terms are added for sim-to-real robustness and smooth control~\cite{Peng_2018,liao2025beyondmimic,chen2025gmt}. The reward encourages the simulated robot to track the converted reference in joint space, root state and key end-effectors, while penalizing stance-foot slip, joint-limit violation and abrupt action changes. The detailed reward definitions and weights are provided in Appendix~\ref{app:tracking_reward}.

\begin{figure}[t]
    \centering
    \includegraphics[width=\textwidth]{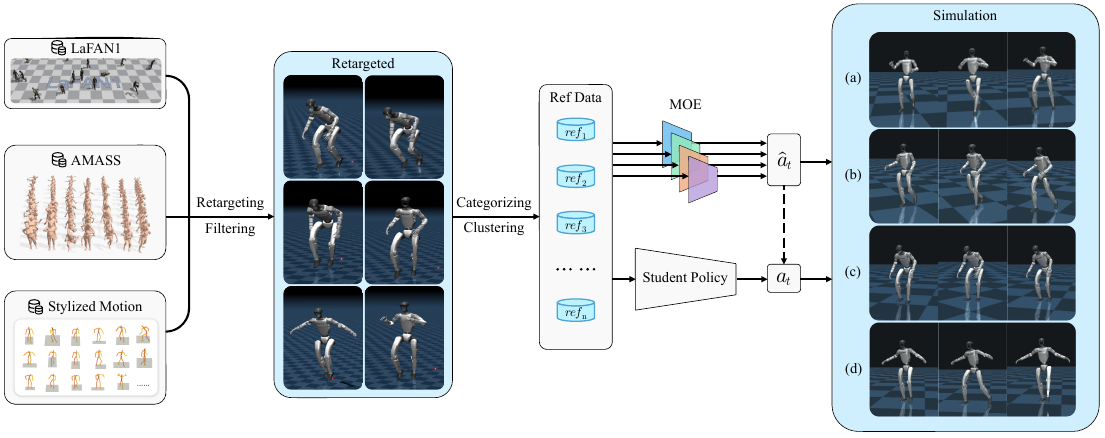}
    \caption{Cluster-and-distill training for stylized whole-body tracking. Expert policies are first trained on content-specific motion clusters with dynamics randomization. Their behaviors are then distilled into a single deployable policy through Dataset Aggregation, allowing the robot to execute diverse stylized references without switching controllers.}
    \label{fig:training_pipeline}
\end{figure}

Because different motion categories exhibit distinct patterns that cannot be well captured by a single policy, the motions are first grouped into four content-level clusters, including straight walking, turning, circular walking and upper-body-emphasized locomotion, and one expert policy is trained for each cluster. During expert training, mild domain randomization is applied to link mass, link inertia, ground friction, joint friction, motor response and sensor noise. The same randomization setting is used for all experts and tracker-ablation variants. After training the experts, DAgger is employed to distill them into a single deployable generalist policy. The whole process is illustrated in Fig.~\ref{fig:training_pipeline}.

\subsection{Deployment Protocol}
The proposed framework is deployed in two stages. In the offline stage, the style-transfer model generates a stylized human-motion reference, which is converted and filtered into a G1-compatible whole-body reference. Diffusion sampling is performed with DDIM using $50$ inference steps. On a workstation with an Intel Core i7-12700 CPU and two NVIDIA GeForce RTX 3080 Ti GPUs, the average pure generation time is about $0.24$ s per action after warm-up. The complete offline export, including generation, saving motion files and video rendering, takes about $0.98$ s per action. Video rendering is used only for visualization and is not part of the robot control loop.

In the online stage, only the reference-preview tracking policy and the low-level PD controller run on the robot. The reference motion is provided at $30$ Hz, the policy is updated at $50$ Hz, and the low-level PD controller runs at $1000$ Hz. Therefore, the real-time requirement is imposed on the tracking policy and the low-level control loop, while diffusion sampling is treated as offline preparation. Hardware trials are terminated when the robot falls, triggers a safety shutdown, or shows unrecoverable tracking divergence.

\section{Experimental Results and Analysis}
\label{sec:experiments}
\subsection{Experimental Platform and Settings}
The experiments evaluate three questions. First, whether the proposed generator can preserve motion content while transferring human-motion-inspired style cues. Second, whether the decoded-motion regularization improves robot-oriented feasibility, especially contact consistency and joint smoothness. Third, whether the converted references can be tracked on a real humanoid robot under different style intensities. Unless otherwise specified, all generated references are processed through the same reference-conversion and tracking pipeline before computing robot-oriented metrics.

\begin{table}[t]
\centering
\caption{Robot deployment and evaluation settings.}
\label{tab:deployment_settings}
\footnotesize
\begin{tabularx}{\textwidth}{@{}lX@{}}
\toprule
\textbf{Item} & \textbf{Setting} \\
\midrule
Robot platform & Unitree G1 humanoid robot \\
Simulation backend & Isaac Sim 5.0 \\
Robot model & Unitree G1, $29$ active DoFs \\
Reference source & Stylized whole-body references generated offline and converted to the G1 robot model \\
Reference conversion & Offline conversion from generated human motion to G1-compatible joint references, with joint-limit handling and infeasible-motion filtering \\
Reference rate & $30$ Hz \\
Policy update rate & $50$ Hz \\
Low-level control rate & $1000$ Hz \\
Control interface & Joint-position targets tracked by low-level PD control \\
PD gains & Joint-specific proportional gains in the range of $50$--$200$ \\
Deployment terrain & Flat indoor ground \\
Trial protocol & Guidance-scale trials over $\lambda \in \{0,0.25,0.5,0.75,1.0\}$, with $25$ hardware trials per scale \\
Success criterion & Completion of the full motion without falling, safety shutdown or unrecoverable tracking divergence \\
Offline computation & DDIM sampling with $50$ inference steps; about $0.24$ s per action for pure generation and about $0.98$ s per action for complete offline export \\
Workstation & Intel Core i7-12700 CPU and $2\times$ NVIDIA GeForce RTX 3080 Ti GPUs \\
\bottomrule
\end{tabularx}
\end{table}

\subsection{Evaluation Metrics for Style and Robot Executability}
We report two groups of metrics. The perceptual group evaluates whether the generated motions remain close to human motion distributions and preserve recognizable content and style. Fréchet Motion Distance (FMD), Content Recognition Accuracy (CRA) and Style Recognition Accuracy (SRA) are computed following common motion style transfer protocols~\cite{Jang_2022,10.1145/3480145,10.1109/TVCG.2023.3320216}. These metrics are useful for comparison with animation-oriented baselines, but they do not directly measure whether a humanoid robot can execute the generated reference.

The robot-oriented group evaluates physical feasibility and closed-loop execution. Foot Sliding Factor (FSF) measures horizontal stance-foot drift per contact frame. A frame is labeled as a contact frame if the foot height is below $h_c=0.05$ m and the inter-frame displacement satisfies
$\|\mathbf{p}_f(t)-\mathbf{p}_f(t-1)\|_2 < 0.02$ m. For foot $f$, the accumulated horizontal sliding is computed as
\begin{equation}
S_f =
\sum_{\substack{t\in\mathcal{C}_f\\ t-1\in\mathcal{C}_f}}
\left\|
\mathbf{p}^{xy}_f(t)-\mathbf{p}^{xy}_f(t-1)
\right\|_2 ,
\label{eq:fsf_sliding}
\end{equation}
where $\mathbf{p}^{xy}_f(t)$ is the horizontal foot position and $\mathcal{C}_f$ is the set of contact frames. FSF is then defined as
\begin{equation}
\mathrm{FSF}=\frac{1}{2}\left(\bar{S}_{L}+\bar{S}_{R}\right),
\label{eq:fsf}
\end{equation}
where $\bar{S}_f=S_f/N_f$ and $N_f=\left|\{t\,|\,t\in\mathcal{C}_f,\ t-1\in\mathcal{C}_f\}\right|$ is the number of valid consecutive contact-frame pairs for foot $f$. A lower FSF indicates more stable stance-foot contact.

$A_{\mathrm{pos}}$ and $J_{\mathrm{pos}}$ measure acceleration-level and jerk-level finite differences of joint positions, reflecting reference smoothness. Given a joint-position trajectory $\mathbf{q}(t)$ with length $T$, they are defined as
\begin{equation}
A_{\mathrm{pos}}=
\frac{1}{T-2}\sum_{t=3}^{T}
\left\|\Delta^2 \mathbf{q}(t)\right\|_1,
\label{eq:apos}
\end{equation}
where
\begin{equation}
\Delta^2 \mathbf{q}(t)=\mathbf{q}(t)-2\mathbf{q}(t-1)+\mathbf{q}(t-2),
\end{equation}
and
\begin{equation}
J_{\mathrm{pos}}=
\frac{1}{T-3}\sum_{t=4}^{T}
\left\|\Delta^3 \mathbf{q}(t)\right\|_1,
\label{eq:jpos}
\end{equation}
where
\begin{equation}
\Delta^3 \mathbf{q}(t)=\mathbf{q}(t)-3\mathbf{q}(t-1)+3\mathbf{q}(t-2)-\mathbf{q}(t-3).
\end{equation}
Lower values indicate smoother references with fewer high-frequency joint oscillations.

Success Rate (SR) is the ratio of trials completed without falling, safety shutdown or unrecoverable tracking divergence. Tracking error is additionally reported in the tracking-strategy ablation to evaluate closed-loop reference-following accuracy. It is computed as the mean normalized joint-pose deviation between the simulated robot state and the converted reference over the controlled joints and rollout horizon.

\subsection{Human Motion Style Generation Results}
Table~\ref{tab:comparison} compares the proposed method with MotionDiffuse~\cite{zhang2022motion}, Motion Puzzle~\cite{Jang_2022} and MCM-LDM~\cite{2024Arbitrary}. The purpose is not to claim the best animation score, but to examine whether style transfer remains executable after conversion to G1-compatible references and tracking. For all methods, the simulated success rate in Table~\ref{tab:comparison} is evaluated under the same reference-conversion and closed-loop tracking protocol. We use $16$ converted actions for each method, and each action is rolled out $256$ times with different random seeds and perturbations, resulting in $4096$ simulation rollouts per method. The reported Sim. SR is computed as the number of successful rollouts divided by $4096$. A rollout is counted as successful if the robot completes the full reference without falling, safety termination or unrecoverable tracking divergence.

MCM-LDM achieves the best FMD and SRA among the reported methods, indicating strong kinematic perceptual quality. However, its higher FSF and lower simulated SR show that perceptual quality alone is insufficient for robot deployment. The proposed method shows a more favorable trade-off, with lower FSF, smoother joint references and a higher simulated closed-loop SR. For MotionDiffuse, FMD, CRA and SRA are not reported because its text-conditioned interface does not provide the same exemplar-style label setting required by the classifier-based style-transfer evaluation protocol; therefore, only robot-oriented feasibility metrics are reported after conversion to G1-compatible references and tracking.

\begin{figure}[t]
    \centering
    \includegraphics[width=1\textwidth]{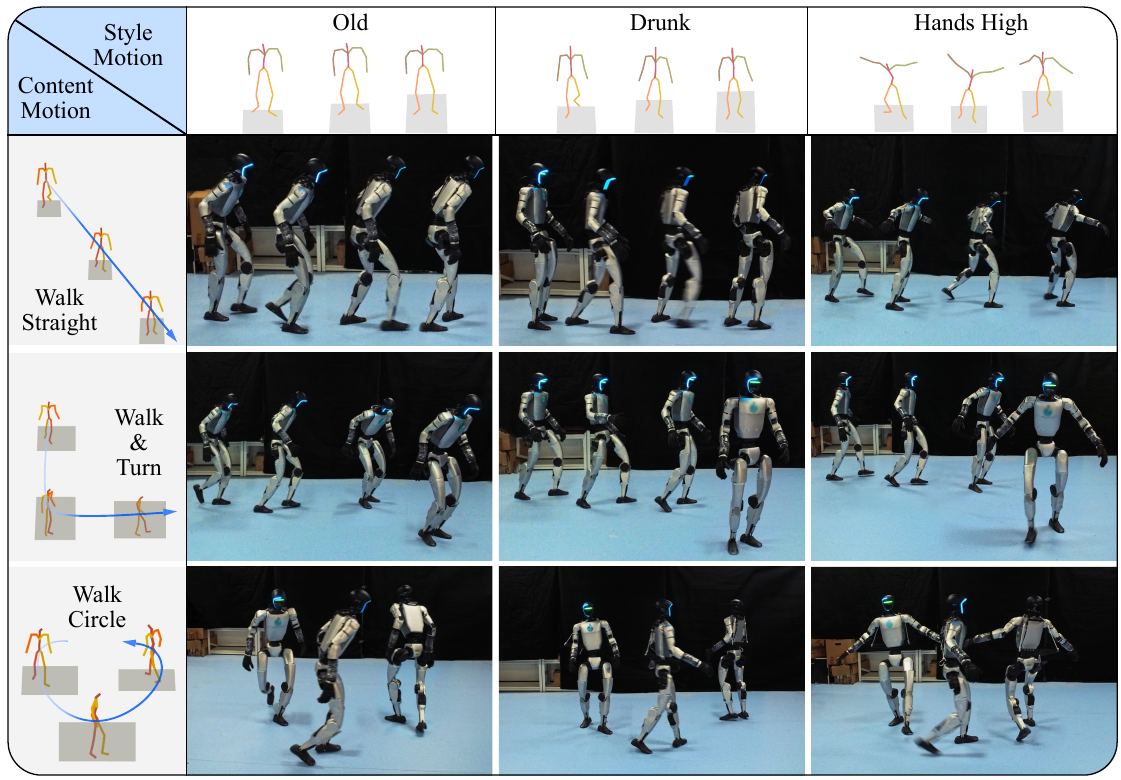}
    \caption{Exemplar-driven style transfer across different contents and styles. Short human style exemplars are transferred to different target content motions while the intended trajectory and action structure are preserved.}
    \label{fig:style_transfer_examples}
\end{figure}

Qualitative examples are shown in Fig.~\ref{fig:style_transfer_examples}. The same human style exemplar can be reused across different content motions, while different exemplars produce visibly distinct gait rhythm, trunk posture and arm-motion patterns.
\begin{table}[ht]
\centering
\caption{Comparison of perceptual quality and robot-oriented physical feasibility. Simulated SR is computed over $4096$ rollouts for each method, using $16$ converted actions and $256$ rollouts per action. $\uparrow$ indicates higher is better, $\downarrow$ indicates lower is better.}
\label{tab:comparison}
\footnotesize
\setlength{\tabcolsep}{2pt}
\begin{tabularx}{\textwidth}{@{}l*{7}{>{\centering\arraybackslash}X}@{}}
\toprule
\multirow{3}{*}{\textbf{Method}} &
\multicolumn{3}{c}{\textbf{Perceptual Quality}} &
\multicolumn{4}{c}{\textbf{Robot-Oriented Feasibility}} \\
\cmidrule(lr){2-4} \cmidrule(lr){5-8}
& \textbf{FMD} $\downarrow$ 
& \textbf{CRA} $\uparrow$ 
& \textbf{SRA} $\uparrow$ 
& \textbf{FSF} $\downarrow$ 
& $\boldsymbol{A_{\mathrm{pos}}}$ $\downarrow$ 
& $\boldsymbol{J_{\mathrm{pos}}}$ $\downarrow$ 
& \textbf{Sim. SR} $\uparrow$ \\
& 
& (\%)
& (\%)
& (m/frame)
& (a.u.)
& (a.u.)
& (\%) \\
\midrule
MotionDiffuse 
& -- & -- & -- 
& 0.006233 & 0.497825 & 0.339368 
& 27.93\% \\

Motion Puzzle 
& 113.31 & 26.31 & 46.33 
& 0.005853 & 0.568928 & 0.822193 
& 67.14\% \\

MCM-LDM 
& \textbf{34.78} & 33.62 & \textbf{58.66} 
& 0.006273 & 0.372012 & 0.316419 
& 59.81\% \\

\textbf{Ours} 
& 39.81 & \textbf{35.18} & 55.55 
& \textbf{0.004722} & \textbf{0.307976} & \textbf{0.283434} 
& \textbf{96.02\%} \\
\bottomrule
\end{tabularx}
\end{table}

\subsection{Ablation of Tracking Strategy}
To evaluate the contribution of the tracking design, we compare four tracker-training strategies under the same set of converted stylized references. The evaluation set in Table~\ref{tab:tracking_strategy_comparison} is different from that in Table~\ref{tab:comparison} and contains more diverse stylized references. This ablation isolates the control stage: the generated references and reference-conversion procedure are fixed, while the tracker training strategy is changed. The evaluation uses $64$ converted stylized references covering different content motions, style exemplars and guidance scales. Each reference is rolled out $64$ times in simulation with different random seeds and perturbations, resulting in $4096$ rollouts for each tracker variant. A rollout is counted as successful if the robot completes the full reference without falling, safety termination or unrecoverable tracking divergence.

Table~\ref{tab:tracking_strategy_comparison} compares four variants. SP-original trains a single policy only on original motion references. SP-stylized additionally includes generated stylized references. SE-only uses clustered specialized experts without distillation. CD-generalist distills the experts into one deployable policy. The results show that stylized references improve adaptation to the generated reference distribution. SE-only achieves the best simulated metrics because each expert specializes in one motion cluster, but it assumes that the correct expert is selected for each reference. CD-generalist slightly sacrifices simulated tracking performance, removes this runtime expert-selection requirement, and is therefore used as the deployed controller.

\begin{table}[t]
\centering
\caption{Ablation of tracker-training strategies under the same converted stylized references. SP denotes a single policy, SE denotes clustered specialized experts, and CD denotes the final cluster-and-distill policy. Simulated success rate is computed over $4096$ rollouts for each variant.}
\label{tab:tracking_strategy_comparison}
\footnotesize
\setlength{\tabcolsep}{3pt}
\begin{tabularx}{\textwidth}{@{}lccccc@{}}
\toprule
\textbf{Strategy} 
& \textbf{Stylized refs.} 
& \textbf{Experts} 
& \textbf{DAgger} 
& \textbf{Tracking err. (a.u.)} $\downarrow$
& \textbf{Sim. SR (count, \%)} $\uparrow$ \\
\midrule
SP-original 
& No & No & No 
& 0.122 & 3328/4096, 81.25\% \\
SP-stylized 
& Yes & No & No 
& 0.096 & 3624/4096, 88.48\% \\
SE-only 
& Yes & Yes & No 
& \textbf{0.083} & \textbf{3856/4096, 94.14\%} \\
\textbf{CD-generalist} 
& Yes & Yes & Yes 
& 0.086 & 3808/4096, 92.97\% \\
\bottomrule
\end{tabularx}
\end{table}

\subsection{Robot-Oriented Physical Feasibility and Generation Ablation}
Table~\ref{tab:feasibility} evaluates the two decoded-motion regularizers used during generation. Contact regularization reduces stance-foot drift, while smoothness regularization reduces high-frequency joint oscillation. The full model obtains the best values across FSF, $A_{\mathrm{pos}}$ and $J_{\mathrm{pos}}$, supporting the role of physics-aware regularization in constraining style expression toward the robot-executable reference space.

These results should be interpreted together with the downstream reference conversion and tracking stages. The decoded-motion regularizers are imposed before conversion to the G1 morphology, and therefore do not mathematically guarantee that contact and smoothness properties are exactly preserved after conversion. Instead, they reduce severe artifacts in the generated human-motion reference. The final executability claim is based on robot-oriented metrics computed after conversion to G1-compatible references and closed-loop tracking.

\begin{table}[h]
\centering
\scriptsize
\caption{Ablation of physics-aware regularization terms.}
\label{tab:feasibility}
\begin{tabular}{lccc}
\toprule
\textbf{Configuration}
& \shortstack[c]{\textbf{FSF} \\ (m/frame)$\downarrow$}
& \shortstack[c]{$A_{\text{pos}}$ \\ (a.u.)$\downarrow$}
& \shortstack[c]{$J_{\text{pos}}$ \\ (a.u.)$\downarrow$} \\
\midrule
w/o both              & 0.005005          & 0.333533          & 0.308064 \\
w/ contact regularization        & 0.004825          & 0.320142          & 0.294568 \\
w/ smoothness regularization     & 0.004770          & 0.319252          & 0.294946 \\
\textbf{w/ both} & \textbf{0.004722} & \textbf{0.307976} & \textbf{0.283434} \\
\bottomrule
\end{tabular}
\end{table}

\subsection{Style Intensity Analysis}
The classifier-free guidance scale $\lambda$ controls the intensity of the transferred style. We evaluate $\lambda \in \{0,0.25,0.5,0.75,1.0\}$ to examine the trade-off between style recognizability and robot executability. As $\lambda$ increases, the generated motion shows stronger style cues, reflected by higher SRA and qualitative body-motion changes. At the same time, excessive style amplification may increase foot sliding, joint jitter and tracking failure. SRA and robot-oriented feasibility metrics are averaged over the generated test motions for each guidance scale, while Hardware SR is computed from $25$ Unitree G1 trials per scale. Table~\ref{tab:lambda_analysis} therefore reports both style and robot-oriented metrics for each guidance scale.

\begin{figure}[t]
    \centering
    \includegraphics[width=0.7\textwidth]{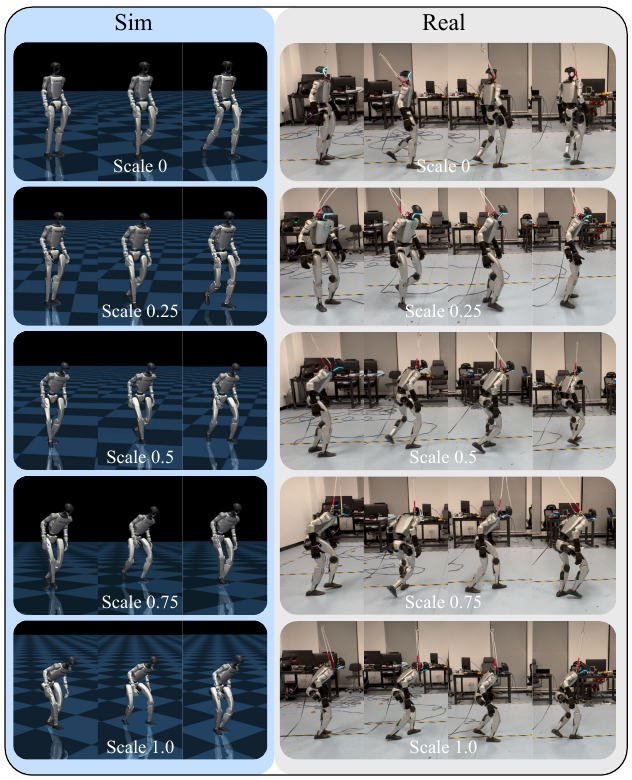}
    \caption{Style intensity control in simulation and real Unitree G1 deployment. Increasing $\lambda$ strengthens human-like expressive cues, such as slower gait rhythm and upper-body sway, while the style intensity must remain within the physically executable range of the humanoid robot.}
    \label{fig:style_intensity_control}
\end{figure}

Figure~\ref{fig:style_intensity_control} visualizes the effect of $\lambda$ in both simulation and real-robot deployment. Larger $\lambda$ values make the transferred style more recognizable, but also increase the risk of contact inconsistency and tracking difficulty.

\begin{table}[t]
\centering
\caption{Effect of guidance scale $\lambda$ on style expression and robot executability. Hardware SR is computed from $25$ Unitree G1 trials per guidance scale.}
\label{tab:lambda_analysis}
\scriptsize
\begin{tabularx}{\textwidth}{@{}cXXXXX@{}}
\toprule
$\boldsymbol{\lambda}$ 
& \textbf{SRA} $\uparrow$ 
& \textbf{FSF} $\downarrow$ 
& $\boldsymbol{A_{\mathrm{pos}}}$ $\downarrow$ 
& $\boldsymbol{J_{\mathrm{pos}}}$ $\downarrow$ 
& \textbf{Hardware SR} $\uparrow$ \\
& (\%)
& (m/frame)
& (a.u.)
& (a.u.)
& (\%) \\
\midrule
0.00 & 17.22 & 0.002125 & 0.106100 & 0.082846 & 100.0\% \\
0.25 & 42.44 & 0.002309 & 0.116894 & 0.092362 & 100.0\% \\
0.50 & 51.56 & 0.002767 & 0.134200 & 0.109798 & 100.0\% \\
0.75 & 55.56 & 0.002877 & 0.135287 & 0.111264 & 96.0\% \\
1.00 & 58.30 & 0.003422 & 0.144717 & 0.120689 & 84.0\% \\
\bottomrule
\end{tabularx}
\end{table}

\subsection{Real-Robot Deployment}
The real-robot study deploys the generated references on a Unitree G1 humanoid robot using the G1-compatible reference conversion and reference-preview tracking policy described in Section~\ref{sec:humanoid_control}. A trial is counted as successful only when the robot completes the full sequence without falling, triggering safety shutdown or showing unrecoverable tracking divergence.

The deployment tests evaluate whether the generated style variations remain trackable after conversion to G1-compatible references. We quantitatively test multiple guidance scales and qualitatively examine representative style-content combinations, and report trial-level success according to the termination criteria in Table~\ref{tab:deployment_settings}. Figure~\ref{fig:style_transfer_examples} shows representative style-content combinations, and Fig.~\ref{fig:style_intensity_control} shows the corresponding intensity modulation in simulation and on hardware.

In the reported guidance-scale experiment, $25$ hardware trials are conducted for each $\lambda \in \{0,0.25,0.5,0.75,1.0\}$, resulting in $125$ Unitree G1 trials. The robot completes $120$ out of $125$ trials, corresponding to a hardware success rate of $96.0\%$. Most failures occur at larger guidance scales, where stronger stylization increases upper-body sway, stance-timing distortion and tracking difficulty.

\subsection{Failure Analysis and Limitations}
Several limitations are observed in deployment. First, excessive style amplification can produce high-frequency jitter and over-extended postures, indicating that the guidance scale must be bounded by robot kinematics and tracking robustness. Second, strong upper-body swing may shift the center of mass and reduce the support margin, especially when the style exemplar contains exaggerated arm motion. Third, asymmetric styles can disturb contact rhythm and cause stance-phase inconsistencies. Fourth, sharp turns and circular paths may increase yaw tracking error because style-dependent trunk motion is coupled with root orientation changes.

A further limitation is possible content-style entanglement. The MotionCLIP style embedding may contain not only expressive cues but also content-related motion tendencies, such as turning bias, stride asymmetry or arm-use patterns. The content and trajectory conditions help preserve the intended action and global path, but they do not provide a formal guarantee of perfect disentanglement. When a style exemplar contains strong turning or asymmetric stepping while the content motion is straight walking, residual yaw bias or stance-timing distortion may appear.

Finally, the current real-robot validation is mainly conducted on flat indoor ground and uses a limited number of hardware trials compared with large-scale simulation evaluation. The present framework does not yet model complex terrain, external contacts or object interaction. Future work will extend the framework to terrain-aware generation, interaction-aware reference conversion and broader hardware evaluation.

\section{Conclusion}
This paper presents a bionic generation-to-control framework for physically executable humanoid motion style transfer. Given a short human style exemplar and a target content motion, the proposed method generates a stylized whole-body reference that preserves the target action while transferring reusable human-motion cues. The generator uses a multi-condition latent diffusion model with explicit condition-token fusion, classifier-free style-intensity control, and decoded-motion contact-consistency and temporal-smoothness regularization. The generated references are converted into $29$-DoF Unitree G1-compatible references and executed by a reference-preview whole-body tracker trained with style-diverse trajectories and a cluster-and-distill strategy.

Experiments demonstrate that the proposed method achieves a favorable balance between style-transfer quality and robot-oriented physical feasibility. Although animation-oriented baselines obtain better kinematic perceptual scores in some reported metrics, the proposed pipeline improves stance-foot stability, joint smoothness and hardware execution success. The reported Unitree G1 experiments achieve a $96.0\%$ success rate over $125$ guidance-scale trials, showing the feasibility of transferring short human style exemplars to real humanoid whole-body motion. Future work will extend the framework to broader real-robot style-content combinations and richer human--robot interaction scenarios.

\appendix
\section{Tracking Reward Design}
\label{app:tracking_reward}

The expert policies are optimized with a weighted reference-tracking reward. To avoid notation conflict with the generation model, the robot joint position is denoted by $\boldsymbol{\theta}_t$, the policy action by $\mathbf{u}_t$, the root pose by $\mathbf{b}_t$, and the root velocity by $\boldsymbol{\nu}_t$. The subscript $\theta$ in the denoising network $\epsilon_{\theta}$ denotes model parameters, whereas the bold symbol $\boldsymbol{\theta}_t$ in this appendix denotes robot joint positions. Superscript $\star$ denotes the converted G1-compatible reference. The total reward is the weighted sum of the terms in Table~\ref{tab:tracking_reward_main}.

\begin{table}[h]
\centering
\caption{Tracking reward terms used for expert-policy training.}
\label{tab:tracking_reward_main}
\footnotesize
\setlength{\tabcolsep}{2pt}
\begin{tabularx}{\textwidth}{@{}l c c >{\raggedright\arraybackslash}X >{\raggedright\arraybackslash}p{0.3\textwidth}@{}}
\toprule
\textbf{Term} & \textbf{Weight} & \textbf{Scale} & \textbf{Definition} & \textbf{Purpose} \\
\midrule
$\rho_{\theta}$ 
& $0.30$ 
& $\kappa_{\theta}=2.0$ 
& $\exp\!\left(-\kappa_{\theta}\|\boldsymbol{\theta}_t-\boldsymbol{\theta}^{\star}_t\|_2^2\right)$ 
& Joint-position tracking over controlled G1 joints. \\

$\rho_{\dot{\theta}}$ 
& $0.08$ 
& $\kappa_{\dot{\theta}}=0.10$ 
& $\exp\!\left(-\kappa_{\dot{\theta}}\|\dot{\boldsymbol{\theta}}_t-\dot{\boldsymbol{\theta}}^{\star}_t\|_2^2\right)$ 
& Joint-velocity tracking to reduce phase lag during style-dependent rhythm changes. \\

$\rho_b$ 
& $0.16$ 
& $\kappa_b=4.0$ 
& $\exp\!\left(-\kappa_b d_b(\mathbf{b}_t,\mathbf{b}^{\star}_t)^2\right)$ 
& Root height, roll and pitch tracking for balance. \\

$\rho_{\nu}$ 
& $0.10$ 
& $\kappa_{\nu}=1.0$ 
& $\exp\!\left(-\kappa_{\nu}\|\boldsymbol{\nu}_t-\boldsymbol{\nu}^{\star}_t\|_2^2\right)$ 
& Root linear velocity and yaw-rate tracking. \\

$\rho_e$ 
& $0.16$ 
& $\kappa_e=8.0$ 
& $\exp\!\left(-\kappa_e |\mathcal{E}|^{-1}\sum_{k\in\mathcal{E}}\|\mathbf{p}_{k,t}-\mathbf{p}^{\star}_{k,t}\|_2^2\right)$ 
& End-effector tracking for feet and hands. \\

$\rho_{\mathrm{slip}}$ 
& $0.06$ 
& $\kappa_{\mathrm{slip}}=25.0$ 
& $\exp\!\left(-\kappa_{\mathrm{slip}}|\mathcal{F}|^{-1}\sum_{f\in\mathcal{F}}m_{f,t}\|\mathbf{v}^{xy}_{f,t}\|_2^2\right)$ 
& Stance-foot slip suppression. \\

$\rho_{\mathrm{lim}}$ 
& $0.05$ 
& $\kappa_{\mathrm{lim}}=10.0$ 
& $\exp\!\left(-\kappa_{\mathrm{lim}}\|[\boldsymbol{\theta}_t-\boldsymbol{\theta}_{\max}]_+ + [\boldsymbol{\theta}_{\min}-\boldsymbol{\theta}_t]_+\|_2^2\right)$ 
& Soft joint-limit regularization. \\

$\rho_u$ 
& $0.06$ 
& $\kappa_u=0.05$ 
& $\exp\!\left(-\kappa_u\|\mathbf{u}_t-\mathbf{u}_{t-1}\|_2^2\right)$ 
& Action-rate smoothing for stable low-level PD tracking. \\

$\rho_{\mathrm{alive}}$ 
& $0.03$ 
& -- 
& $1$ if no termination condition is triggered, otherwise $0$ 
& Sequence completion under fall and safety checks. \\
\bottomrule
\end{tabularx}
\end{table}

The reward used at control step $t$ is therefore computed as the sum of each term multiplied by its weight in Table~\ref{tab:tracking_reward_main}. The weights are kept fixed for all expert policies and all tracker-ablation variants.

\begin{table}[h]
\centering
\caption{Symbols used in the tracking reward.}
\label{tab:tracking_reward_symbols}
\footnotesize
\begin{tabularx}{\textwidth}{@{}lX@{}}
\toprule
\textbf{Symbol} & \textbf{Meaning} \\
\midrule
$\boldsymbol{\theta}_t$ & Controlled G1 joint positions at control step $t$. \\
$\dot{\boldsymbol{\theta}}_t$ & Controlled G1 joint velocities. \\
$\mathbf{u}_t$ & Policy action, represented as joint-position targets. \\
$\mathbf{b}_t$ & Root pose components used for tracking, including root height, roll and pitch. \\
$\boldsymbol{\nu}_t$ & Root velocity components, including local linear velocity and yaw angular velocity. \\
$\mathbf{p}_{k,t}$ & Cartesian position of end-effector $k$. \\
$\mathcal{E}$ & End-effector set used for tracking, including feet and hands. \\
$\mathcal{F}$ & Foot set used for slip regularization. \\
$m_{f,t}$ & Stance mask of foot $f$ at step $t$. \\
$\mathbf{v}^{xy}_{f,t}$ & Horizontal velocity of foot $f$. \\
$\boldsymbol{\theta}_{\min},\boldsymbol{\theta}_{\max}$ & Lower and upper joint limits of the G1 model. \\
$(\cdot)^{\star}$ & Converted G1-compatible reference quantity. \\
$[\cdot]_+$ & Element-wise positive part. \\
$d_b(\cdot,\cdot)$ & Root-pose distance over height, roll and pitch. \\
$\kappa_{\cdot}$ & Exponential penalty scale. \\
\bottomrule
\end{tabularx}
\end{table}

\backmatter

\bmhead{Author Contributions}

T.H. and M.Z. contributed equally to this work, including conceptualization, methodology, implementation, experiments, data analysis and writing the original draft. Y.G., F.Y., J.G., X.Z., D.Z. and S.Y. contributed to experiments, data processing, validation and manuscript revision. Y.W., W.G. and S.Z. supervised the work and contributed to review and editing. All authors read and approved the final manuscript.

\bmhead{Acknowledgements}

The authors have no additional acknowledgements to declare.

\bmhead{Funding}

This work was supported in part by the National Natural Science Foundation of China under Grants U22B2040, in part by the Major Project of Anhui Province's Science and Technology Innovation Breakthrough Plan (202423h08050003), in part by the Key Science \& Technology Project of Anhui Province (202523j08050001), and in part by the Fundamental Research Funds for the Central Universities with grant No. YD2090002019.

\bmhead{Data Availability}

The data that support the findings of this study are available from the corresponding author upon reasonable request.

\bmhead{Conflict of Interest} The authors certify that there is no conflict of interest pertaining to the content of this paper.

\bmhead{Ethical Approval and Consent to Participate} Not applicable.

\bmhead{Consent for Publication} Not applicable.
\bibliography{sn-bibliography}

\end{document}